\documentclass{article}

 \usepackage[preprint]{neurips_2026}

\usepackage[utf8]{inputenc} %
\usepackage[T1]{fontenc}    %
\usepackage{hyperref}       %
\usepackage{url}            %
\usepackage{booktabs}       %
\usepackage{multirow}       %
\usepackage{amsmath,amsfonts}       %
\usepackage{nicefrac}       %
\usepackage{microtype}      %
\usepackage{xcolor}         %
\usepackage{graphicx}         %

\usepackage{titlesec}
\titlespacing*{\paragraph}{0pt}{0.5ex plus 0.3ex minus 0.1ex}{1em}

\title{Tying the Loop - Tied Expert Layers in \\Mixture-of-Experts Language Models}

\author{%
  Martin Jaggi\\
  EPFL \\
}

\begin{document}

\maketitle

\begin{abstract}
Mixture-of-Experts (MoE) architectures efficiently scale Large Language Models (LLMs) by activating only a small fraction of their experts per token, yet the full parameter count—dominated by the expert parameters—must be held in training and inference memory.
To address this, we introduce Expert Tying, an architectural modification that shares expert parameters across consecutive transformer layers while preserving independent, layer-wise routing and attention.
We evaluate this approach across common, state-of-the-art architectures, including OLMoE, Qwen3, and DeepSeek-style MoEs.
Our pretraining experiments demonstrate that tying experts can reduce memory footprint by almost $2\times$ at virtually no degradation in perplexity or downstream quality.
By exploiting the parameter redundancy inherent in MoE pathways, our method provides a highly favorable compute-to-memory trade-off, advancing efficient training and scaling of next-generation LLMs.\\~\\
Our codebase is public at \href{https://github.com/epfml/looped-moe}{github.com/epfml/looped-moe}
\end{abstract}

\section{Introduction}
\label{sec:intro} 

Mixture-of-Experts (MoE) architectures have become a standard technique for scaling language models: by activating only a small subset of expert feed-forward networks (FFNs) per token, they decouple total parameter count from per-token compute \cite{shazeer2017outrageously,fedus2022switch}. Recent open-weights MoE models push this decoupling to extremes. DeepSeek-V3 activates only 37B of its 671B parameters per token ($\approx\!5.5\%$), Qwen3-235B-A22B $\approx\!9.4\%$, and Kimi-K2 $\approx\!3.2\%$. As this active fraction shrinks, the memory footprint of an MoE is governed almost entirely by parameters that sit idle in any given forward pass: the full model must reside in training and inference memory even though only a tiny percentage of parameters contribute compute for each token.

This is in strong tension with a second trend: reasoning models and looped-depth models aim to extract more compute from each unique parameter, building more capable models at the same parameter count---the kind of parameter efficiency recently incentivized by OpenAI's Parameter Golf challenge \cite{parametergolf2026}. From this second perspective, standard MoE looks like a step backward---inflating memory with parameters that are often inactive.

\paragraph{Our approach.} We propose \emph{expert tying} to reconcile the two opposing trends of the compute vs.\ memory trade-off. By reusing the same expert FFN weights across a group of consecutive layers while keeping routers, attention, and normalization layer-specific, we preserve MoE's low per-token compute yet raise compute per unique parameter, removing the memory penalty of sparsity rather than the sparsity itself.
Concretely, given a group of $g$ layers, the gate/up/down projections of the $N$ experts are aliased across all $g$ layers, reducing the unique FFN parameters by a factor of $g$; each layer still computes its own routing distribution and its own attention output, so the hidden state continues to flow through $g$ distinct layer operators, not $g$ copies of the same one. The implementation is simple---a single Python-level pointer assignment in HuggingFace transformers models---and requires no changes to training or inference infrastructure beyond the optimizer, which must correctly accumulate gradients from the tied parameters' multiple use sites.

\paragraph{Why this works.} The intuition is that the FFN expert pool can be shared across nearby layers because per-layer attention keeps each layer's effective operator distinct: even with identical expert parameters, attending over a different mixture of token positions produces a different transformation at each depth. Our component ablation (Section~\ref{sec:component-ablation}) confirms this directly: tying the router across consecutive layers barely affects loss, while tying attention costs an order of magnitude more. Expert tying therefore trades a small amount of expressive capacity for a large parameter reduction, with attention---not routing---doing the work of keeping layers distinct. Our experiments confirm this scales: across three state-of-the-art MoE architectures (OLMoE, Qwen3-MoE, DeepSeekMoE) and both fine-grained and coarse-grained expert configurations, group sizes up to $g=4$ yield minimal degradation in validation loss and downstream accuracy, while cutting total FFN parameters by~75\%.

\paragraph{Relation to looped transformers.} Expert tying is a \emph{partial loop}: the MoE FFN sub-block is shared (``looped'') across $g$ consecutive layers, while the rest of the block is not. This places it in the growing family of looped/recurrent-depth designs \cite{dehghani2019universal,geiping2025huginn,zhu2025ouro,prairie2026parcae}, most of which loop the entire block on dense models; by sharing only the FFN sub-block in a feedforward (non-recurrent) MoE stack, we capture the parameter-efficiency benefit of block reuse at the compute and memory profile of an ordinary MoE. A separate line of MoE-specific work shares parameters across layers but each commits to one component---routers, expert pools, or whole blocks \cite{gu2026pathmoe,tan2025rexmoe,csordas2024moeut,chen2026moue}---without isolating which one keeps tied layers distinct; Megrez2 \cite{li2025megrez2} productionizes a single configuration like ours. Our contribution is not the configuration itself but the controlled study identifying, across architectures and granularities, which components can be shared at little quality cost.

\paragraph{Contributions.} This paper makes the following contributions:
\begin{enumerate}
\item \textbf{Expert tying works at scale.} On three common MoE architectures (OLMoE, Qwen3-MoE, DeepSeekMoE) with up to 7B parameters, tying expert FFN weights in groups of four reduces total parameter count by up to $52\%$ with negligible degradation in pretraining loss or downstream accuracy, and delivers a wall-clock training speed-up of up to $23.7\%$.
 
\item \textbf{Per-layer attention, not routing, is what differentiates layers.} 
A controlled component ablation shows that untying attention across tied layers improves loss by an order of magnitude more than untying only the router, even though freed routers visibly diversify their expert choices. This explains why expert tying is cheap and identifies tying experts (largest parameter pool, modest cost) as the highest-leverage component to share.  We further identify that first and last layers should not be looped or tied --- a $2{+}2$ untied prelude and coda yields the single largest architectural gain in our ablation.
 
\item \textbf{Heterogeneous width expansion} reinvests the parameters saved by expert tying into more experts in the tied middle layers, consistently outperforming the untied baseline at iso-parameter count on all three architectures.
This recasts expert tying as a depth-vs-width design axis rather than just a compression technique.

\item We show that expert tying \textbf{composes cleanly with standard MoE training} recipes (load balancing, Muon/AdamW optimization), and requires no architectural changes to attention or routing, making it a drop-in modification for existing MoE codebases.
\end{enumerate}

\section{Related Work}
 
\subsection{Mixture-of-Experts Architectures}
 
Mixture-of-Experts (MoE) architectures decouple total parameter count from per-token compute by activating only a subset of expert FFN networks per token \cite{shazeer2017outrageously,lepikhin2021gshard,fedus2022switch}.
Recent work has pushed toward \emph{fine-grained} experts with large expert pools: DeepSeek-V3 \cite{deepseekv3} uses 256 routed experts with an auxiliary-loss-free load-balancing strategy; Qwen3 \cite{yang2025qwen3} employs 128 experts; Kimi-K2 scales to 384. These models demonstrate that richer expert combinations with smaller individual experts consistently improve expressiveness and downstream quality. OLMoE \cite{muennighoff2024olmoe} shows that even at the 7B-total / 1B-active scale, MoE can outperform dense models trained with 6--7$\times$ more compute. Routing mechanisms remain an active research area: ReMoE \cite{wang2025remoe} replaces top-$k$ + softmax routing with continuous ReLU routing for full differentiability, and Expert Choice routing \cite{zhou2022expertchoice} inverts the standard token-to-expert assignment.
 
Despite strong scaling properties, MoE models carry a substantial memory footprint due to the large number of expert parameters. While only a small fraction of the parameters are activated per token, the full parameters still need to reside in memory. This motivates cross-layer expert sharing as a means to reduce the unique parameter count without changing the activated compute budget.

\subsection{Looped Transformers and Parameter Sharing}
 
Our work is closely connected to a growing body of research on \emph{looped transformers}---architectures that reuse the same block of layers multiple times, increasing effective depth without proportionally increasing parameter count. Under this lens, our approach is a partial loop: the MoE FFN sub-block is shared (``looped'') across a group of consecutive layers, while attention, normalization, and routers remain distinct at each position in the group.
 
\paragraph{Foundations and theory.}
Universal Transformers \cite{dehghani2019universal} first demonstrated that sharing a single transformer block across depth can match standard transformers on certain tasks via adaptive-computation-time halting. \cite{giannou2023looped} provide a constructive expressivity result, showing that a constant-depth transformer placed in a loop can emulate a programmable instruction-set computer.
\cite{saunshi2025latent} formalized the reasoning inductive bias of this design, showing both theoretically and empirically that a $k$-layer transformer looped $L$ times closely matches the performance of a $kL$-layer non-looped model on reasoning-intensive tasks, despite using $L\times$ fewer unique parameters. Their central claim---that many reasoning problems require depth but not necessarily parameters---provides the theoretical basis for the broader looped transformer program.
 
\paragraph{Scaling looped pretraining.}
Recent work has demonstrated that looped architectures can be scaled effectively. \emph{Huginn} \cite{geiping2025huginn} introduces a depth-recurrent 3.5B-parameter LM with a prelude--core--coda topology, trained on 800B tokens, where unrolling the core block more times at inference improves downstream quality. \emph{Ouro} \cite{zhu2025ouro} pretrains 1.4B and 2.6B looped models for 7.7T tokens, matching the performance of 4B and 8B standard transformers respectively. Ouro's controlled experiments offer a key insight for our work: looping does not increase knowledge \emph{capacity} per parameter (both looped and non-looped models encode roughly 2 bits/parameter on synthetic memorization tasks), but it preserves or improves knowledge \emph{manipulation} on multi-hop reasoning. \emph{Parcae} \cite{prairie2026parcae} establishes the first scaling laws for looped architectures, showing that a 770M Parcae matches a 1.3B parameter transformer on the same data, and that compute-optimal training requires scaling loop count and data together. Parcae also identifies stability bottlenecks (residual explosion, loss spikes) specific to loop-based training and addresses them via spectral-norm constraints on the input injection.
Most recently, \cite{schwethelm2026recurrence} fit scaling laws for fully-looped language models and find that each additional recurrence contributes substantially less than an additional unique block, providing a quantitative complement to our component-ablation perspective.
 
\paragraph{Differentiating iterations.}
A central challenge in fully-looped architectures is that the same block processes every iteration, limiting per-depth specialization. Several methods add lightweight per-iteration adaptation on top of a shared core. \emph{Relaxed Recursive Transformers} \cite{bae2025relaxed} attach depth-wise LoRA modules to each loop iteration of a dense recursive transformer, recovering most of the performance of the original full-size model when retrofitting pretrained LLMs into looped form. \emph{RingFormer} \cite{heo2025ringformer} generates input-dependent level signals via depth-specific low-rank matrices that modulate the shared block at each iteration. \emph{LoopFormer} \cite{jeddi2026loopformer} trains a looped transformer on variable unrolling depths with a consistency loss between short and long trajectories.
\emph{MeSH} \cite{yu2025mesh} externalizes state management into an explicit memory buffer with step-wise routers to diversify computation across iterations. \emph{Mixture-of-Recursions} \cite{bae2025mor} uses lightweight routers to assign different recursion depths to individual tokens, unifying parameter sharing with adaptive computation. \emph{ModernALBERT / Mixture of LoRAs} \cite{nouriborji2025mol} inserts LoRA experts with token-conditional routing inside the shared FFN of a recursive encoder. \emph{SpiralFormer} \cite{yu2026spiralformer} introduces multi-resolution recursion, operating the shared core at varying sequence resolutions across loop iterations. All of these works operate on dense (non-MoE) transformers.
Concurrently, \emph{Hyperloop Transformers} \cite{zeitoun2026hyperloop} loop a middle block (with untied begin/end layers) in dense transformers, adding loop-level hyper-connections to differentiate iterations%
. Their dense design needs an explicit per-iteration mechanism; our MoE design achieves the same differentiation by keeping attention per-layer while tying only experts.

\subsection{Cross-Layer Sharing in MoEs}
 
The idea of sharing expert parameters across layers in MoE architectures has emerged in several recent works. Because the FFN/expert block carries the bulk of parameters in a modern MoE, sharing experts is the most impactful axis of parameter reduction---and, through the lens of looping, corresponds to looping only the MoE sub-block while leaving attention and routing free to vary per layer.
 
\emph{MoEUT} \cite{csordas2024moeut} combines fully-shared MoE layers in a Universal Transformer framework. MoEUT uses $\sigma$-MoE for both attention and feedforward layers, groups non-shared layers into blocks that are then recurrently stacked, and introduces a custom ``peri-layernorm'' scheme. MoEUT shares every component---including routers and attention---across loop iterations, and does not explore which components to untie.
 
\emph{Megrez2} \cite{li2025megrez2} is architecturally closest to our work: it partitions an $L$-layer transformer into groups of $n$ consecutive layers that share expert weights while retaining per-layer gating networks, attention, and normalization. Combined with pre-gated routing for memory-efficient expert prefetching, Megrez2 achieves competitive performance with 7.5B total / 3B active parameters. Megrez2 is a single production system optimised for inference efficiency on memory-constrained devices, and reports one tying configuration without isolating its contribution to model quality. Our work gives a controlled decomposition that varies group size, tying topology, expert granularity, and base architecture (OLMoE, Qwen3-MoE, DeepSeekMoE), and that isolates \emph{which} component of a tied MoE layer is responsible for cross-depth differentiation.
 
The looped MoE variant of \cite{chen2026loopmoe} is concurrent work on the special case of fully-looped MoE blocks, sharing all components across iterations and then restoring per-iteration distinctness with an architecture modification, an adaptive-layernorm modulation conditioned on the iteration index. Our finding is more direct: the component ablation shows that per-iteration distinctness is carried by attention, which improves over just distinct layernorm gains between loops.%
 
\emph{ReXMoE} \cite{tan2025rexmoe} approaches cross-layer expert reuse from a complementary angle. Rather than tying expert weights, ReXMoE expands each layer's \emph{candidate expert pool} to include experts from adjacent layers, with per-layer routers selecting from a union of its own and its neighbors' experts. This increases routing diversity without adding parameters, using a progressive scaling routing (PSR) strategy during training. Our work is complementary: we share the actual weight tensors (reducing unique parameters) while keeping routers per-layer, whereas ReXMoE keeps weights independent while broadening routing scope.
 
\emph{PathMoE} \cite{gu2026pathmoe} shares \emph{router} parameters (rather than expert weights) across consecutive layers, arguing that independent per-layer routing over $N$ experts across $L$ layers creates a too large path space of $N^L$ possible expert sequences, while tokens in practice concentrate on a small fraction of coherent linguistic paths. Shared routers constrain the effective path space and improve sample efficiency at 0.9B and 16B scale. PathMoE is the mirror image of our approach---they share routers with distinct experts, while we share experts with distinct routers. A related line of work coordinates routing across layers without sharing any weights: \cite{hu2026synergistic} adds an auxiliary cross-layer coupling loss that aligns top-$k$ routing across adjacent layers, improving expert specialization; this is complementary to weight tying/looping and further supports that cross-layer routing structure is a meaningful axis.
 
\emph{Mixture of Universal Experts} (MoUE) \cite{chen2026moue} takes the sharing idea to its extreme: a single expert pool is shared across \emph{all} layers of the network. To make this work, MoUE introduces a staggered rotational topology for structured expert exposure, a universal load-balancing loss that accounts for repeated expert access across depth, and a universal router with trajectory state to coordinate routing across layers. Where MoUE unifies the expert pool globally, our work studies the intermediate regime of group-wise sharing, which preserves more per-layer flexibility at the cost of less aggressive parameter reduction. Our ablations over group size effectively trace the continuum between per-layer independent experts ($g=1$) and fully shared experts ($g=L$, MoUE-like).
 
\emph{WideNet} \cite{xue2022widenet} is an earlier dense approach that shares both FFN and self-attention weights across all transformer layers. Unlike our work, WideNet is dense (non-MoE) and ties all components.

\section{Which Components Should Be Tied Across Layers?}
\label{sec:component-ablation}

A standard MoE transformer layer comprises four learned components: the
expert FFN weights, the self-attention Q,K,V,O projections, the router, and
the normalization layer gains (the RMSNorm scaling parameters). Each can in principle be
tied (or ``looped'') across a group of consecutive layers or kept independent. Before
turning to large-scale experiments on production MoE architectures
(Section~\ref{sec:results}), we establish on a smaller controlled
architecture which of these components must remain unique-per-layer to
preserve quality, and which can share weights cheaply.

\paragraph{Setup.} A depth-$32$ decoder-only transformer with
$d_\text{model}=512$, $n_\text{heads}=8$, sequence length~$512$, and a
fine-grained MoE FFN ($32$ experts, top-$k=8$, $d_\text{ff,expert}=128$).
All blocks are pre-norm~\cite{xiong2020layer} with RMSNorm
\cite{zhang2019rmsnorm}, rotary position embeddings (RoPE)
\cite{su2024roformer}, and QK-norm \cite{henry2020query} applied to
queries and keys before the dot product. Full architectural and
optimisation details are in Appendix~\ref{app:repro-ablation}.
We sweep three orthogonal axes:
\begin{itemize}
\item \textbf{Tying mode.} The three modes \textsc{all-tie},
\textsc{attn-tie}, and \textsc{expert-tie} differ in which of the
four layer components are tied within a group; they form a monotone
chain from most to least tied (Table~\ref{tab:modes}).
\item \textbf{Topology.} The $28$ middle layers form $k$ tying groups,
each consisting of a parameter block of $b$ unique transformer layers
reused $\ell$ times in succession (so $k \cdot b \cdot \ell = 28$); we
denote this~$k\,(\text{group of }b)^{\ell\times}$. The $2$-layer prelude
and $2$-layer coda are always left untied. We sweep five topologies:
$1\,(\text{group of }2)^{14\times}$,
$2\,(\text{group of }2)^{7\times}$,
$3\,(\text{group of }2)^{(5,4,5)\times}$ (non-uniform loop count,
$5{+}4{+}5{=}14$),
$4\,(\text{group of }1)^{7\times}$, and
$7\,(\text{group of }1)^{4\times}$. As an additional reference point
we evaluate a universal transformer (MoEUT) \cite{csordas2024moeut} configuration in which
all $32$ layers form a single tying group with no untied prelude or coda
($1\,(\text{group of }2)^{16\times}$ over the full stack).
\item \textbf{Granularity.} Fine ($32$ experts, top-$8$) versus coarse
($8$ experts, top-$2$, with per-expert FFN width scaled up so that active
compute is unchanged).
\end{itemize}
All ablation runs use the same loss recipe as the production-architecture
experiments in Section~\ref{sec:results}: a load-balancing auxiliary loss
\cite{shazeer2017outrageously, fedus2022switch} at coefficient
$\alpha_{\text{aux}} = 10^{-2}$ and a router $z$-loss~\cite{zoph2022stmoe} at coefficient
$\alpha_z = 10^{-4}$, both added to the next-token cross-entropy.

\begin{table}[h]
\centering
\small
\caption{Tying modes. Within a tying group of $g$ consecutive layers, each component is either \emph{tied} (one parameter tensor shared by all $g$ layers) or \emph{per-layer} (each layer has its own copy). Norm gains are always per-layer in our codebase (each transformer layer has its own RMSNorm).}
\label{tab:modes}
\begin{tabular}{lcccc}
\toprule
Mode & FFN experts & Attention & Router & Norm gains \\
\midrule
\textsc{all-tie}     & tied & tied & tied        & per-layer \\
\textsc{attn-tie}    & tied & tied & per-layer   & per-layer \\
\textsc{expert-tie}  & tied & per-layer & per-layer & per-layer \\
\bottomrule
\end{tabular}
\end{table}

\paragraph{Routing diagnostic.} Beyond final validation loss, we report
\emph{cross-loop agreement}: the average fraction of routing decisions
that match across pairs of layers within the same tying group\footnote{%
For each token we compare the top-$1$ (\texttt{argmax}) expert
selected by each layer in a tying group, and report the fraction of
layer pairs that select the \emph{same} top-$1$ expert, averaged over
tokens and over the first $3$ validation batches. The metric reflects
top-$1$ routing only, irrespective of lower-ranked expert choices overlap.
}, measured
on the held-out validation set at the end of training. A value of $1$
means tied layers always pick the same experts for the same tokens; a
value of $0$ means they pick disjoint expert subsets. The metric isolates
the per-layer differentiation produced by the router, irrespective of
whether the underlying expert weights or attention projections are also
tied, and is well-defined for all three modes (in \textsc{all-tie} mode,
the same router still sees different hidden states at each layer, so
agreement is non-trivial). Cross-loop agreement therefore lets us decouple
the question ``do tied layers \emph{behave} like one operator?'' from
the question of model quality (loss).

\begin{table}[t]
\centering
\caption{Fine-grained ablation ($32$ experts, top-$k=8$): final validation
loss at $10{,}000$ steps and cross-loop agreement. ``$\Delta$'' is loss
relative to the MoE baseline (no cross-layer tying). ``Total'' is unique
parameter count; ``Saved'' is the relative reduction vs.\ the baseline.
``MoEUT'' denotes a single $32$-layer tying group with no prelude or coda;
all other rows have $28$ middle layers in $k$ tying groups with a $2{+}2$
untied prelude/coda. Random chance cross-loop agreement (32 experts) is 0.031.
Within each topology, modes are listed in order of
increasing per-layer freedom.}
\label{tab:component-ablation}
\begin{tabular}{lrrrrr}
\toprule
Configuration & Total~~ & Saved $\uparrow$ & Loss $\downarrow$ & $\Delta$ $\downarrow$ & \!\!\!\!Cross-loop\\
& params & memory &&& \!\!\!\!agreement \\
\midrule
\emph{Baseline (no cross-layer tying)} & & & & & \\
\quad MoE, no tying                                  & $488$M & ---      & 3.432 & 0.000  & ---   \\
\quad dense, no tying                                & 186M   & $61.9\%$ & 3.560 & +0.128 & ---   \\

\midrule
\emph{MoEUT-style ($1\,(\text{group of }2)^{16\times}$, no prelude/coda)}\!\!\!\!\!\!\!\!\!\!\!\!\!\!\! & & & & & \\
\quad MoEUT \textsc{all-tie}                         & $65$M  & $86.7\%$ & 3.650 & +0.218 & 0.740 \\
\quad MoEUT \textsc{attn-tie}                        & $66$M  & $86.6\%$ & 3.653 & +0.221 & 0.071 \\
\midrule
\emph{Mid-stack ($28$ middle layers, $2{+}2$ prelude/coda)}\!\!\!\!\!\!\!\! & & & & & \\
\quad $1\,(\text{group of }2)^{14\times}$, \qquad~\textsc{all-tie}         & $121$M & $75.2\%$ & 3.553 & +0.121 & 0.785 \\
\quad $1\,(\text{group of }2)^{14\times}$, \qquad~\textsc{attn-tie}        & $121$M & $75.1\%$ & 3.548 & +0.116 & 0.155 \\
\quad $1\,(\text{group of }2)^{14\times}$, \qquad~\textsc{expert-tie}      & $148$M & $69.6\%$ & 3.490 & +0.058 & 0.087 \\

\quad $2\,(\text{group of }2)^{7\times}$, \qquad~~\textsc{all-tie}          & $161$M & $67.1\%$ & 3.511 & +0.079 & 0.733 \\
\quad $2\,(\text{group of }2)^{7\times}$, \qquad~~\textsc{attn-tie}         & $161$M & $67.0\%$ & 3.512 & +0.080 & 0.160 \\
\quad $2\,(\text{group of }2)^{7\times}$, \qquad~~\textsc{expert-tie}       & $186$M & $61.9\%$ & 3.469 & +0.037 & 0.121 \\

\quad $3\,(\text{group of }2)^{(5,4,5)\times}$, ~~\textsc{all-tie}    & $151$M & $69.1\%$ & 3.481 & +0.049 & 0.756 \\
\quad $3\,(\text{group of }2)^{(5,4,5)\times}$, ~~\textsc{attn-tie}   & $151$M & $69.0\%$ & 3.481 & +0.049 & 0.193 \\
\quad $3\,(\text{group of }2)^{(5,4,5)\times}$, ~~\textsc{expert-tie} & $174$M & $64.5\%$ & 3.451 & +0.019 & 0.107 \\

\quad $4\,(\text{group of }1)^{7\times}$, \qquad~~\textsc{all-tie}          & $161$M & $67.1\%$ & 3.525 & +0.093 & 0.831 \\
\quad $4\,(\text{group of }1)^{7\times}$, \qquad~~\textsc{attn-tie}         & $161$M & $67.0\%$ & 3.516 & +0.084 & 0.126 \\
\quad $4\,(\text{group of }1)^{7\times}$, \qquad~~\textsc{expert-tie}       & $186$M & $61.9\%$ & 3.470 & +0.038 & 0.090 \\

\quad $7\,(\text{group of }1)^{4\times}$, \qquad~~\textsc{all-tie}          & $202$M & $58.6\%$ & 3.485 & +0.053 & 0.764 \\
\quad $7\,(\text{group of }1)^{4\times}$, \qquad~~\textsc{attn-tie}         & $202$M & $58.5\%$ & 3.485 & +0.053 & 0.142 \\
\quad $7\,(\text{group of }1)^{4\times}$, \qquad~~\textsc{expert-tie}       & $224$M & $54.1\%$ & 3.454 & +0.022 & 0.126 \\
\bottomrule
\end{tabular}
\end{table}

\subsection{Finding 1: Per-layer attention drives loss}
\label{sec:finding-attention}

Table~\ref{tab:component-ablation} reveals a sharp decomposition.
At fixed topology, untying \emph{only} the router
(\textsc{all-tie}~$\to$~\textsc{attn-tie}) cuts cross-loop agreement
by $\sim 4{-}7\times$ at every mid-stack topology
($0.733 \to 0.160$ at $2$-group, $0.831 \to 0.126$ at $4$-group), and by
$\sim 10\times$ at MoEUT ($0.740 \to 0.071$). Untying \emph{also} the
attention (\textsc{attn-tie}~$\to$~\textsc{expert-tie}) produces only a
residual $\sim 1.2{-}1.8\times$ further drop. Per-layer routers do strongly
individuate layers when free to, even though the underlying expert
weights remain identical. The loss response is the inverse:
\begin{itemize}
\item Untying the router (\textsc{all-tie}~$\to$~\textsc{attn-tie})
changes loss by at most $0.009$ in any cell, and the sign is not
consistent across topologies (\textsc{attn-tie} better at $1$- and
$4$-group; \textsc{all-tie} marginally better at MoEUT and $2$-group;
tied at $3$- and $7$-group). Within the resolution of this ablation, the
loss effect of router-untying is essentially zero.
\item Untying attention (\textsc{attn-tie}~$\to$~\textsc{expert-tie})
reduces loss by $0.030$--$0.058$ at every topology---several
times the router effect, consistent in sign and magnitude across
runs.
\end{itemize}

The two effects are decoupled: routing diversity is produced almost
entirely by the router, while the loss benefit of distinct per-layer
operators flows almost entirely through attention. Per-layer routing
freedom is observable in the routing distribution but does not translate
into model capability when the attention operator is shared across
the group.

\paragraph{Component cost ranking.}
Reading the chain at $2$-group fine yields
$\Delta(\text{tie experts}) \approx +0.037$,
$\Delta(\text{also tie attention}) \approx +0.043$,
$\Delta(\text{also tie router}) \approx 0$ (slightly negative,
$-0.001$), summing to the cumulative
gap baseline\,$\to$\,\textsc{all-tie} of $0.079$. The MoEUT-style
configurations (which tie all components together) correspond to the
regime measured at scale by \cite{schwethelm2026recurrence}; their
finding that full-block recurrence carries substantial capacity cost
is consistent with the gap we observe.

Two practical conclusions follow. First, expert-tying and
attention-tying carry comparable single-component costs
($\approx +0.037$ and $\approx +0.043$ at $2$-group fine), but
the FFN expert pool dominates attention by an order of magnitude in
parameter count, so expert-tying remains the highest-leverage memory
move per parameter saved. Second, the router can be tied at no measurable
cost on top of expert-and-attention-tying. The $+0.043$ attention figure
is its cost \emph{conditional on} experts already being tied; we did
not run an attention-only-tied configuration. The ranking motivates
our main design (Section~\ref{sec:results}): tie expert weights, keep
attention per-layer, and treat the router as tied or untied indifferently.

\subsection{Finding 2: Topology choice within \textsc{expert-tie} gives major memory savings, without compromising quality}
\label{sec:finding-topology}

Once the tying mode is fixed at \textsc{expert-tie}, partitioning the
$28$ middle layers into groups is a relatively minor knob: loss across
$k \in \{2, 3, 4, 7\}$ groups varies by only $0.019$ ($3.451$ to $3.470$),
and adding $1$-group (the most aggressive sharing, at $148$M unique params)
extends the range to $0.039$. Cross-loop agreement is similarly stable
(range $0.087$--$0.126$), confirming that topology choice does not
qualitatively change routing behaviour. Topology within \textsc{expert-tie}
therefore sets the parameter-savings ratio without large quality
consequences---a knob we use in Section~\ref{sec:results} to interpret
tie-group size as a parameter-efficiency dial.

\subsection{Finding 3: Component ranking holds for all expert granularities}
\label{sec:finding-granularity}

A natural concern is that the dominance of attention over the router
might depend on the routing capacity afforded by fine-grained MoE: with
$32$ experts and top-$k=8$, each layer's router has rich combinatorial
flexibility; with coarse routing ($8$ experts, top-$2$) it does not.
We re-run the MoE baseline, MoEUT \textsc{all-tie}, MoEUT \textsc{attn-tie},
the three-mode $2$-group cross-section, and three \textsc{expert-tie}
topologies at the coarse setting, with per-expert FFN width scaled up
so active FFN compute is unchanged.
\begin{table}[t]
\centering
\caption{Coarse-grained variant ($8$ experts, top-$k=2$, active FFN
compute matched). The component ordering and the loss-vs-routing
decoupling observed at fine granularity hold without modification.
``Total'' is unique parameter count; ``Saved'' is the relative reduction
vs.\ the coarse baseline. 
Random chance cross-loop agreement (8 experts) is 0.125.}
\label{tab:granularity}
\begin{tabular}{lrrrrr}
\toprule
Configuration & Total~~ & Saved $\uparrow$ & Loss $\downarrow$ & $\Delta$ $\downarrow$ & \!\!\!\!Cross-loop\\
& params & memory &&& \!\!\!\!agreement \\
\midrule
\emph{Baseline MoE (no cross-layer tying)} & $488$M & ---     & 3.481 & 0.000 & ---    \\
\midrule
MoEUT \textsc{all-tie} ($1\,(\text{group of }2)^{16\times}$, full stack)\!\!\!\!\!\!\!\!  & $65$M  & $86.7\%$ & 3.663 & +0.182 & 0.896  \\
MoEUT \textsc{attn-tie} ($1\,(\text{group of }2)^{16\times}$, full stack)\!\!\!\!\!\!\!\! & $65$M  & $86.7\%$ & 3.664 & +0.183 & 0.172  \\
\midrule
$2\,(\text{group of }2)^{7\times}$, \textsc{expert-tie} & $186$M & $61.9\%$ & 3.489 & +0.008 & 0.192  \\
$4\,(\text{group of }1)^{7\times}$, \textsc{all-tie}    & $161$M & $67.1\%$ & 3.559 & +0.078 & 0.847  \\
$4\,(\text{group of }1)^{7\times}$, \textsc{attn-tie}   & $161$M & $67.0\%$ & 3.546 & +0.065 & 0.226  \\
$4\,(\text{group of }1)^{7\times}$, \textsc{expert-tie} & $186$M & $61.9\%$ & 3.489 & +0.008 & 0.185  \\
$7\,(\text{group of }1)^{4\times}$, \textsc{expert-tie} & $224$M & $54.2\%$ & 3.475 & $-0.006$ & 0.185  \\
\bottomrule
\end{tabular}\vspace{-2mm}
\end{table}
\\
The pattern in Table~\ref{tab:granularity} is the same as at fine expert
granularity. The router-untying step (\textsc{all-tie}~$\to$~\textsc{attn-tie}
at $4$-group) collapses cross-loop agreement by $3.7\times$ ($0.847 \to
0.226$) while moving loss by only $0.013$. The attention-untying step
(\textsc{attn-tie}~$\to$~\textsc{expert-tie}) produces a $1.22\times$ residual
agreement drop and a $0.057$ loss improvement. Topology spread within
\textsc{expert-tie} remains small ($0.014$). The component ranking, the
routing--loss decoupling, and the practical conclusion are robust to
granularity.

\subsection{Finding 4: Untied prelude/coda dominates all other architectural choices}
\label{sec:finding-prelude}

The largest single effect in either Table~\ref{tab:component-ablation}
or Table~\ref{tab:granularity} is the gap between MoEUT-style
configurations and any mid-stack configuration. At fine granularity,
the gap spans $0.097$--$0.202$, dwarfing the within-mid-stack mode
effect ($\sim 0.05$), the topology spread within \textsc{expert-tie}
($\sim 0.04$), and the granularity effect. At coarse, the analogous
gap is $0.104$--$0.189$.

The table provides a clean mode-matched isolation: MoEUT
\textsc{all-tie} ($3.650$) versus $1$-group mid-stack \textsc{all-tie}
($3.553$) differ by $0.097$, and the corresponding \textsc{attn-tie}
pair differs by $0.105$. The two configurations differ only in the
size of the tying region ($32$ vs.\ $28$ middle layers) and in the
presence of an untied $2{+}2$ prelude/coda; since topology width
within \textsc{expert-tie} accounts for at most $0.04$, the bulk of
the $\sim 0.10$ mode-matched gap is attributable to the prelude/coda
itself. The interpretation is that the first and last layers of
the stack are qualitatively different from the rest---they perform
embedding-input and lm-head-output adjustments---and forcing them
into the shared parameter pool of a fully-recurrent design imposes
a penalty that no expressive routing or attention freedom in the
middle can recover. Practitioners adopting expert tying should
preserve at least a thin untied prelude and coda.

\subsection{Optimization dynamics of tied experts}
\label{sec:opttied}
When sharing expert parameters across $g$ consecutive layers, the
tied-parameter learning rate must be scaled to control the per-step
update magnitude. The classical heuristic for shared weights
\cite{lecun1998efficient,hoffer2017train} prescribes $1/\sqrt{g}$
under the random-walk assumption, but Muon's Newton--Schulz
orthogonalisation normalises the backward-pass gradient magnitude
regardless of $g$, so the relevant argument concerns
\emph{forward-pass} amplification: the same $\Delta W$ is applied
$g$ times in sequence within the forward pass. A linear penalty
($1/g$) corresponds to the conservative case in which a token routes to the
same expert at every depth, producing residual-stream growth
analogous to the explosion observed in fully looped architectures
\cite{prairie2026parcae}. With independent per-layer routers and
intermediate self-attention, tokens take diverse, shifting paths and
amplification compounds sub-linearly, motivating a milder $1/\sqrt{g}$
rule.

\paragraph{Empirical validation.}
We ablate the divisor on the $4$-group topology. Averaging the final validation loss across all three tying modes (\textsc{all-tie}, \textsc{attn-tie}, \textsc{expert-tie}), the $1/\sqrt{g}$ rule ($3.503$) and the strict $1/g$ rule ($3.506$) perform near-identically,
while \emph{no} scaling is clearly worst ($3.542$). This suggests that forward-pass amplification compounds sub-linearly under independent routing and attention. We adopt $1/\sqrt{g}$ as the default for the remainder of this paper: it matches the linear rule's empirical performance while reflecting
the milder amplification expected when tokens take diverse paths.

\section{Main Experiments on Production MoE Architectures}
\label{sec:results}

The component ablation of Section~\ref{sec:component-ablation} converged
on a clear design recipe: tie expert FFN weights, keep attention and
routers per-layer, and preserve a $2{+}2$ untied prelude and coda. We
now apply this recipe to three production MoE architectures
(OLMoE \cite{muennighoff2024olmoe}, Qwen3-MoE \cite{yang2025qwen3},
DeepSeekMoE \cite{deepseekv3}) and study two practical questions:
(i) how much parameter saving does $g{=}4$ expert tying buy, and at what
quality cost? (ii) does heterogeneous \emph{width expansion}---reinvesting
the parameters saved by tying as additional experts in the tied middle
layers---recover quality at iso-parameter count, turning expert tying
from compression into a depth-vs-width design axis?

\paragraph{Setup.} For each architecture we train five configurations.
The \emph{baseline} ($g{=}1$) is the standard untied MoE; two \emph{tied}
configurations share expert weights across consecutive groups of $g{=}2$
and $g{=}4$ layers, with a $2{+}2$ untied prelude and coda; and two
\emph{width-expanded} configurations widen the tied middle layers (at
$g{=}4$) to $2\times$ and $4\times$ the baseline expert count, with the
$4\times$ variant chosen so that total parameter count returns to within
$1\%$ of the baseline (``iso-base''). Active parameters per forward pass
are identical within each architecture across all configurations. All
variants train on a $75{:}25$ mixture of DCLM-edu \cite{dclm-edu} and
FinePhrase \cite{finephrase} for $20{,}000$ steps at effective batch
size $524$k tokens ($\approx 10.5$B tokens total). Models are
scaled-down variants of the named production architectures; full
hyperparameters in Appendix~\ref{app:repro-main}.

\paragraph{Optimizer.} We use Muon \cite{jordan2024muon, liu2025muon}
for the $2$D hidden weights---attention Q,K,V,O projections and FFN
gate/up/down projections---at base learning rate $\eta_\text{Muon}
= 2 \times 10^{-2}$ with weight decay $0.1$. AdamW
\cite{loshchilov2019adamw} handles the embeddings, output head, norm
gains, biases, and routers at $\eta_\text{AdamW} = 0.1 \cdot
\eta_\text{Muon} = 2 \times 10^{-3}$ with weight decay $0.01$.\footnote{%
Routers are kept on AdamW even though their weights are $2$D, because
the router output behaves as a per-token classifier head where adaptive
per-parameter learning rates suit the heavy-tailed gradient distribution.
Weight decay for AdamW parameters according to \cite{jordan2024muon}.}
Tied expert weight tensors receive an LR scaled by $1/\sqrt{g}$
(Section~\ref{sec:opttied}); without this the tied stack effectively
trains at a higher step size on its expert weights, since gradients
accumulate from $g$ layer use-sites into the same parameter. Weight
decay is left uncompensated: tied and untied parameter groups use the
same base $\lambda$ (baseline $\lambda = 0.1$ on Muon, $\lambda = 0.01$
on AdamW; see Appendix~\ref{app:repro-main}). Both optimisers follow the same cosine
schedule (linear warmup, decay to $0.1\times$ peak). All configurations train with a load-balancing auxiliary loss
\cite{shazeer2017outrageously, fedus2022switch} at coefficient
$\alpha_{\text{aux}} = 10^{-2}$ and a router $z$-loss~\cite{zoph2022stmoe} at coefficient
$\alpha_z = 10^{-4}$, both added to the next-token cross-entropy. We extensively monitor router health and confirm that cross-layer tying does not induce expert collapse (see Appendix~\ref{app:router-health} for detailed routing dynamics). Full details and hyperparameters are given in Appendix~\ref{app:repro-main}.

\begin{table}[htbp]
\centering
\small
\caption{Final pretraining and downstream metrics on three MoE
architectures (native depths: 16 layers for OLMoE/DeepSeekMoE, 28 for Qwen3-MoE), $20{,}000$ steps ($\approx 10.5$B tokens). Per-architecture
configurations: \emph{baseline} = standard untied MoE; \emph{$g{=}4$} =
expert weights tied across consecutive groups of $4$ layers with $2{+}2$
untied prelude/coda; \emph{$+\,w{\times}$ width} = the tied middle layers
widened to $w$ times the baseline expert count.
The $4\times$ row matches
each baseline's total-parameter budget (``iso-base''). \emph{Active} is
parameters used per token in the forward pass; it is unchanged across
configurations within each architecture, since neither cross-layer tying
nor width expansion changes top-$k$ compute. \emph{Saved} is the relative
reduction in total parameters vs.\ that architecture's baseline. LM Loss on validation set, not including aux losses.
\emph{Avg Acc} is the macro-average $3$-shot accuracy on
\{ARC-Easy, ARC-Challenge, HellaSwag, PIQA, WinoGrande, OpenBookQA\} via
\texttt{lm-evaluation-harness}. Arrows mark direction of preference.}
\label{tab:tiny_20k_final}
\begin{tabular}{llrrrrrr}
\toprule
Arch & Topology & Active\footnote{%
Active parameters are counted as forward-pass FLOPs in parameter-equivalents.
In contrast, 'Total' reflects physical (unique) parameter storage. If different experts are routed to at each layer, the two ways of counting are identical.
However, if the same expert is selected multiple times within a tied group of layers, each layer's computation contributes to the active parameter count, even though no new unique weights are loaded into memory.%
} & Total & Saved $\uparrow$ & Loss $\downarrow$ & PPL $\downarrow$ & \!\!\!Avg Acc\% $\uparrow$ \\
\midrule
\multirow{5}{*}{DeepSeekMoE}
& baseline                     & \multirow{5}{*}{$158$M} & $523$M & ---           & $3.132$ & $22.92$ & $41.5$ \\
& $g{=}2$                      &                          & $372$M & $29\%$        & $3.149$ & $23.30$ & $41.2$ \\
& $g{=}4$                      &                          & $296$M & $43\%$        & $3.180$ & $24.06$ & $41.0$ \\
& $g{=}4$, $2{\times}$ width   &                          & $372$M & $29\%$        & $3.143 $ & $23.19$ & $40.9$ \\
& $g{=}4$, $4{\times}$ width \emph{(iso-base)}\!\!\!\!\!\! & & $523$M & $\approx 0\%$ & $3.105$ & $22.31$ & $41.9$ \\
\midrule
\multirow{5}{*}{OLMoE}
& baseline                     & \multirow{5}{*}{$170$M} & $523$M & ---           & $3.119$ & $22.62$ & $41.9$ \\
& $g{=}2$                      &                          & $372$M & $29\%$        & $3.136$ & $23.00$ & $42.1$ \\
& $g{=}4$                      &                          & $296$M & $43\%$        & $3.171$ & $23.83$ & $40.8$ \\ 
& $g{=}4$, $2{\times}$ width   &                          & $372$M & $29\%$        & $3.132$ & $22.93$ & $41.5$ \\
& $g{=}4$, $4{\times}$ width \emph{(iso-base)}\!\!\!\!\!\! & & $523$M & $\approx 0\%$ & $3.089$ & $21.95$ & $42.1$ \\
\midrule
\multirow{5}{*}{Qwen3-MoE}
& baseline                     & \multirow{5}{*}{$112$M} & $459$M & ---           & $3.171$ & $23.84$ & $41.7$ \\
& $g{=}2$                      &                          & $300$M & $35\%$        & $3.201$ & $24.55$ & $40.6$ \\
& $g{=}4$                      &                          & $220$M & $52\%$        & $3.238$ & $25.47$ & $39.8$ \\
& $g{=}4$, $2{\times}$ width   &                          & $300$M & $35\%$        & $3.192$ & $24.34$ & $40.4$ \\ 
& $g{=}4$, $4{\times}$ width \emph{(iso-base)}\!\!\!\!\!\! & & $459$M & $\approx 0\%$ & $3.142$ & $23.17$ & $41.6$ \\
\bottomrule
\end{tabular}
\end{table}

\paragraph{Expert tying $g{=}4$ is essentially free.}
Going from baseline to tied/looped configurations saves $29$--$52\%$ of total
parameters at a modest quality cost that scales monotonically with the
tying group size. At $g{=}2$ the loss penalty is only $0.02$--$0.03$
(OLMoE $3.119{\to}3.136$, DeepSeekMoE $3.132{\to}3.149$, Qwen3-MoE
$3.171{\to}3.201$); at $g{=}4$ it grows to $0.05$--$0.07$
(OLMoE $+0.052$, DeepSeekMoE $+0.048$, Qwen3-MoE $+0.067$) while saving
$43$--$52\%$ of total parameters. Average downstream accuracy follows the
same trend, dropping by at most $1.9\%$ at $g{=}4$. The cost is
consistent across all three architectures and free of any cliff: the
smooth baseline${\to}\,g{=}2\,{\to}\,g{=}4$ progression shows that quality
degrades gracefully as more layers share weights. We'll show below that larger models show even less degradation when tying experts.

\paragraph{At iso-parameter count, width expansion beats the untied baseline.}
Reinvesting the parameters saved by tying as additional experts in the tied
middle layers (the $g{=}4$, $4\times$ width variant) returns to the baseline's
parameter budget---and consistently \emph{exceeds} it. Loss improves by
$\approx 0.03$ on all three architectures (OLMoE $3.089$ vs $3.119$,
DeepSeekMoE $3.105$ vs $3.132$, Qwen3-MoE $3.142$ vs $3.171$) with downstream
accuracy matched or better, a gain consistent in sign and magnitude across
architectures. The effect persists at full scale: the $7$B $g{=}4$, $2\times$ width
model improves on the untied baseline in both loss ($2.812$ vs $2.820$) and
accuracy ($48.2\%$ vs $47.4\%$; Table~\ref{tab:fullscale}). At a fixed
parameter budget, spending capacity on \emph{wider experts shared across tied
layers} is thus a better use of parameters than a standard untied stack.

\paragraph{Efficiency and throughput gains.}
Beyond memory savings, expert tying accelerates wall-clock training. With
fewer unique parameter tensors the architecture incurs less weight-loading
bandwidth, smaller optimizer state updates, and reduced gradient
communication under data parallelism. In our setup ($4{\times}$H200 GPUs,
PyTorch DDP), the $g{=}4$ tied large OLMoE configuration sustains $51{,}777$
tokens/sec---a $23.7\%$ speed-up over the untied baseline ($41{,}859$
tokens/sec). On the smaller config shown in Table \ref{tab:tiny_20k_final}, the throughput gain is $15.7\%$ (again with DDP on 4 GPUs). Throughput gains are increasing further for large models. %
See Appendix~\ref{app:repro-main} for detailed measurements.

\paragraph{Scaling to $7$B model size.}
To confirm expert tying holds beyond our reduced-scale sweep, we train the \emph{full-size} OLMoE-1B-7B architecture ($7.12$B total, $\approx\!1.3$B active; $16$ layers, $2048$ hidden, $64$ experts, top-$8$) from scratch, in three configurations: untied baseline, $g{=}4$ tying, and $g{=}4$ with $2\times$ width, each for $30{,}000$ steps ($\approx\!15.7$B tokens) on $4\times$H200
GPUs under the same recipe.%
The result mirrors the small scale and is in fact stronger: $g{=}4$ tying
\emph{matches the untied baseline} in both loss ($2.825$ vs $2.820$) and
downstream accuracy ($47.5\%$ vs $47.4\%$) while using \emph{half} the total
parameters ($3.50$B vs $7.12$B, active compute unchanged). %
Reinvesting part of the saving as $2\times$-width experts ($4.71$B total) \emph{exceeds} the baseline even more significantly in loss, PPL and downstream accuracy. 
As an external reference point, our $g{=}4$ configuration also surpasses the official OLMoE-1B-7B-0924 checkpoint (\texttt{step5000-tokens20B}, $44.4\%$ at $\approx\!20$B tokens) even when trained on only half the tokens, and even with a $g{=}4$ looped model only half the size---though this is a reference point rather than a controlled comparison, as that model differs in tokenizer, data, and optimizer. 
Expert tying thus preserves quality at billion-parameter
scale while halving the memory footprint.

\begin{table}[htbp]
\centering
\small
\caption{Full-scale OLMoE-1B-7B ($7.12$B total, $\approx 1.3$B active)
under expert tying, $30{,}000$ steps ($\approx 15.7$B tokens). Columns as
in Table~\ref{tab:tiny_20k_final}. 
Compared to the baseline, the $g{=}4$ looped model reaches the same downstream accuracy with less than half the model parameters, and at $23.7\%$ higher throughput (Full details in Appendix~\ref{app:repro-main}). Width expansion increases accuracy further (while inherently saving slightly less parameters).
Our model also beats the official OLMoE-1B-7B-0924
\texttt{step5000-tokens20B} checkpoint evaluated under the same protocol. 
While our $g{=}4$ looped model reached $45.1\%$ after $\approx\!10.5B$ tokens ($20{,}000$ steps), the official OLMoE-1B-7B-0924 reached $44.4\%$ average accuracy only after $\approx\!20B$ tokens, despite having about twice as many model parameters. (Note though that the official OLMoE-1B-7B-0924 \texttt{step5000-tokens20B} training recipe differs from our recipe in tokenizer ($50$K vs our $100$K vocab), training data, and optimizer (AdamW vs Muon)).
 }
\label{tab:fullscale}
\begin{tabular}{lrrrrr}
\toprule
Configuration & Total & Saved $\uparrow$ & Loss $\downarrow$ & PPL $\downarrow$ & \!\!\!Avg Acc\% $\uparrow$ \\
\midrule
baseline ($g{=}1$)            & $7.12$B & ---      & $2.82$ & $16.78$ & $47.4$ \\
$g{=}4$                       & $3.50$B & $50.9\%$ & $2.82$ & $16.86$ & $47.5$ \\
$g{=}4$, $2{\times}$ width    & $4.71$B & $33.8\%$ & $2.81$ & $16.64$ & $48.2$ \\
\midrule
\emph{OLMoE-1B-7B-0924} \emph{(ref., $\approx\!20B$ tok)}\!\!\! & $6.9$B\footnote{%
OLMoE-1B-7B-0924 model architecture and parameter count is identical to our baseline, except the slight difference in embedding size due to increased vocabulary size of the cl100k tokenizer in all our runs.} & --- & N/A & N/A & $44.4$ \\
\bottomrule
\end{tabular}
\end{table}

\section{Conclusion}

MoE lowers compute per unique parameter, leaving most weights idle and the model memory-bound, whereas reasoning models raise this ratio. Expert tying reconciles the two: sharing expert FFN weights across consecutive layers preserves the low per-token compute of MoE yet raises compute per unique parameter, removing the memory cost of sparsity rather than the sparsity itself. 
A controlled ablation shows that per-layer attention, not routing, is what keeps tied layers distinct, so the largest parameter pool is the cheapest to share. Out change not only saves parameters but also improves throughput and reduces communication cost.
Reinvested as additional experts, the saved parameters render width and depth exchangeable at a fixed parameter budget.

\paragraph{Limitations.} Our experiments reach 7B parameters at a fixed token budget, leaving frontier scale and longer-horizon training untested. Width expansion is competitive rather than uniformly dominant. Our implementation uses PyTorch without tied-layer-aware kernels, so the reported efficiency gains are a lower bound.

\begin{ack}
We thank Vinko Sabolčec for pointing out grouped GEMM, and to the EPFL RCP compute cluster team for the infrastructure.
MJ acknowledges funding from SNSF Grant 10005248, from the Swiss AI Initiative Projects a139 and a140, from EU Horizon ELSA, from Google, Huawei and Microsoft Lingua.
\end{ack}

\begin{small}
\bibliographystyle{alpha}
\bibliography{references}
\end{small}

\appendix
\newpage

\section{Reproducibility details: Component ablations (Section~\ref{sec:component-ablation})}
\label{app:repro-ablation}

This appendix documents the full hyperparameter configuration used in the
component-tying ablation of Section~\ref{sec:component-ablation},
sufficient to reproduce all reported numbers. The accompanying codebase
will be released on acceptance.

\paragraph{Architecture.} We use a vanilla depth-32 decoder-only transformer with
$d_\text{model}=512$, $n_\text{heads}=8$, $d_\text{head}=64$, sequence
length $512$, and a tiktoken \texttt{cl100k\_base} tokenizer (vocab size
$100{,}277$). Each transformer layer is pre-norm \cite{xiong2020layer}
with RMSNorm \cite{zhang2019rmsnorm}: a multi-head self-attention
sub-block---using rotary position embeddings (RoPE) \cite{su2024roformer}
and QK-norm \cite{henry2020query} (RMSNorm applied to queries and keys
before the RoPE rotation and the dot product)---followed by a
sparsely-activated MoE FFN sub-block with SwiGLU activations
\cite{shazeer2020glu}. The fine-grained MoE uses $32$ experts per layer
with top-$k=8$ and $d_\text{ff,expert}=128$; the coarse-grained variant
uses $8$ experts with top-$k=2$ and $d_\text{ff,expert}=256$, holding
active FFN compute approximately constant. All linear projections
(attention Q,K,V,O, FFN gate/up/down, router) have biases disabled.
Every input sequence starts with a \texttt{<BoD>} token, allowing to channel attention-sink behavior.

\paragraph{Tying topologies.} The depth-32 stack is partitioned into an
untied 2-layer prelude, $k$ tying groups covering the 28 middle layers,
and an untied 2-layer coda. Each tying group is a parameter block of $b$
unique transformer layers reused $\ell$ times consecutively; the group
therefore spans $b \cdot \ell$ layers, and the topology overall covers
$k \cdot b \cdot \ell = 28$ middle layers. We denote a topology with $k$
groups, block size $b$, and loop count $\ell$ as
$k\,(\text{group of }b)^{\ell\times}$. The five topologies we sweep are:
$1\,(\text{group of }2)^{14\times}$,
$2\,(\text{group of }2)^{7\times}$,
$3\,(\text{group of }2)^{(5,4,5)\times}$ (non-uniform loop count
$5{+}4{+}5{=}14$),
$4\,(\text{group of }1)^{7\times}$, and
$7\,(\text{group of }1)^{4\times}$.
The MoE baseline corresponds to no cross-layer tying (all 32 layers
carry independent parameters). We additionally evaluate a MoEUT-style
topology \cite{csordas2024moeut} in which the entire 32-layer stack
forms a single tying group ($1\,(\text{group of }2)^{16\times}$) with
no untied prelude or coda; this serves as the maximal-tying extreme of
the design space.

\paragraph{Tying modes.} Within each group, the chosen subset of components
is tied across the group's layers (i.e., each layer in the group references
the same parameter tensor for the tied components, while the layers compute
forward on distinct activations as in any standard transformer stack).
Norm gains are always per-layer in our codebase (each transformer
layer has its own RMSNorm scaling parameter) and are therefore not
listed as a tying option. We evaluate three modes (Table~\ref{tab:modes}):
\begin{itemize}
\item \textsc{all-tie}: FFN expert weights, attention Q,K,V,O projections,
and the router are all tied across the group.
\item \textsc{attn-tie}: FFN expert weights and attention Q,K,V,O projections
are tied; the router is per-layer.
\item \textsc{expert-tie}: only the FFN expert weights are tied; attention
and router are per-layer.
\end{itemize}
Gradient accumulation through tied parameters' multiple use sites is
handled natively by PyTorch autograd; no manual gradient manipulation
is required.

\paragraph{Optimizer.} Following the established recipe of
\cite{jordan2024muon, liu2025muon}, we use Muon for all $2$D hidden
weights (attention Q,K,V,O projections, FFN gate/up/down projections) with
$\eta_\text{Muon}=2 \times 10^{-2}$, momentum $0.95$ with Nesterov,
weight decay $0.1$, and $5$ Newton--Schulz iterations for the
orthogonalisation step. The remaining parameters (token and output
embeddings, norm gains, router weights, biases) are optimised with
AdamW \cite{loshchilov2019adamw} at $\eta_\text{AdamW}=2 \times 10^{-3}$
(i.e., $0.1 \times \eta_\text{Muon}$, the empirical ratio used by
\cite{jordan2024muon}), $\beta_1=0.9$, $\beta_2=0.95$, and zero weight
decay. Routers are kept on AdamW because their output behaves as a
per-token classifier head, where adaptive per-parameter learning rates
suit the sparse and heavy-tailed gradient distribution.\footnote{%
This differs from the convention in some Muon-MoE implementations
\cite{liu2025muon} that route $2$D router weights through Muon. Our
preliminary experiments found neither convention substantially superior
at the scales considered here; we use the AdamW-router convention
uniformly across all experiments in this paper.} The ablation
configurations in this section never use cross-layer tying that affects
$2$D hidden weights touched by Muon (FFN expert weights are $3$D and
handled separately), so no tied-LR scaling is applied here; the
production-architecture experiments in Section~\ref{sec:results} do
introduce such scaling and are documented in Appendix~\ref{app:repro-main}.

\paragraph{Learning rate schedule.} Both optimizers follow a cosine
schedule with $50$-step linear warmup, decaying from peak to
$\eta_\text{min} = 0.1 \times \eta_\text{peak}$ at the final step.
The Muon and AdamW peaks decay synchronously.

\paragraph{Loss.} Standard next-token cross-entropy plus a load-balancing
auxiliary loss \cite{shazeer2017outrageously,fedus2022switch} with
coefficient $\alpha_\text{aux} = 10^{-2}$ and a router $z$-loss with
coefficient $\alpha_z = 10^{-4}$. The same loss recipe is used in the
production-architecture experiments of Section~\ref{sec:results}
(Appendix~\ref{app:repro-main}).

\paragraph{Training.} Effective batch size $32 \times 8 \times 512
= 131{,}072$ tokens per optimization step (per-step batch $\times$ gradient
accumulation $\times$ sequence length), trained for $10{,}000$ steps
($\approx 1.3$B tokens total). Gradient clipping at global norm $1.0$.
Mixed precision: bfloat16 forward and backward with float32 master weights.
We enable \texttt{torch.compile}.

\paragraph{Data.} A $75{:}25$ mixture of DCLM-edu \cite{dclm-edu} and FinePhrase
\cite{finephrase}, streamed from HuggingFace with shuffle buffers of $10{,}000$
documents per source and a fixed seed (42) for reproducibility.
A held-out 10M-token validation slice (sampled deterministically by skipping
the first 10M tokens of the stream before evaluation) is used for all
loss reporting.

\paragraph{Random seeds.} Model initialization, data shuffling, and stochastic
optimizer behavior are all seeded (default seed $42$). Within-sweep variance
from CUDA matmul nondeterminism is empirically below $0.005$ in final loss
across re-runs, much smaller than the architectural effects under study.

\paragraph{Downstream evaluation.} We report only validation loss for the
ablations of Section~\ref{sec:component-ablation}. The relevant signal
in this section is the relative comparison between tying modes at fixed
parameter budget; absolute downstream task accuracy at this model scale
is below the resolution at which our task suite reliably distinguishes
architectural variants. Downstream numbers for our main configurations,
evaluated at larger model scale, are reported in Section~\ref{sec:results}
(Table~\ref{tab:tiny_20k_final}).

\paragraph{Hardware.} Each ablation run completes in approximately $3$ hours on a single H100 GPU.

\section{Reproducibility details: Main experiments (Section~\ref{sec:results})}
\label{app:repro-main}

This appendix documents the configuration of the production-architecture
runs in Table~\ref{tab:tiny_20k_final}, sufficient to reproduce the
reported numbers given the released codebase.

\paragraph{Architectures.} Scaled-down %
variants of OLMoE
\cite{muennighoff2024olmoe}, Qwen3-MoE \cite{yang2025qwen3}, and
DeepSeekMoE \cite{deepseekv3}, instantiated through the HuggingFace
\texttt{transformers} v5 reference implementations. All three are
pre-norm \cite{xiong2020layer} with RMSNorm \cite{zhang2019rmsnorm},
SwiGLU FFN activations~\cite{shazeer2020glu}, and rotary position
embeddings \cite{su2024roformer}; \emph{Qwen3-MoE} additionally applies
QK-norm~\cite{henry2020query} (RMSNorm on queries and keys) by default,
whereas \emph{OLMoE} and \emph{DeepSeekMoE} use the QKV-clipping
mechanism inherited from the OLMoE config (clip threshold $8$) instead. 
Our DeepSeekMoE-style configuration adopts DeepSeek's fine-grained routing
(top-$6$ over $64$ experts, without using shared experts). It does \emph{not} use Multi-head Latent Attention (MLA), as it is only Transformers v5.9 and does not support grouped GEMM. Since expert tying operates only on the FFN sub-block and routing, it is independent of the attention variant and is expected to compose well with any alternative attention mechanism, as for example MLA.

\emph{OLMoE} and \emph{DeepSeekMoE}: $d_\text{model}=512$,
$16$ layers, $4$ attention heads (no GQA), $64$ routed experts of FFN
intermediate size $256$ each, top-$k=8$ for OLMoE and top-$k=6$ for
DeepSeekMoE; the only difference between the two is top-$k$, which
affects active compute but not total parameter count. \emph{Qwen3-MoE}:
$d_\text{model}=384$, $28$ layers, $6$ attention heads with grouped-query
attention (1 KV head), $60$ routed experts of MoE intermediate size $192$
each, top-$k=4$. All three architectures use the HuggingFace tiktoken
\texttt{cl100k\_base} tokenizer (vocab size $100{,}277$) with
$\text{tie\_word\_embeddings}=\text{False}$. 
Same as in our smaller ablations setup, every input sequence starts with a \texttt{<BoD>} token, allowing to channel attention-sink behavior.

Active and total parameter
counts per configuration are reported in Table~\ref{tab:tiny_20k_final}.

\paragraph{Tying topology and width expansion.} For each architecture
the five configurations are: \emph{baseline} ($g{=}1$, no tying);
\emph{$g{=}2$} and \emph{$g{=}4$} (expert FFN tensors aliased across
consecutive groups of $2$ or $4$ middle layers, $2$-layer prelude and
$2$-layer coda untied);
\emph{$g{=}4$, $2{\times}$ width} and \emph{$g{=}4$, $4{\times}$ width};
the latter two double or quadruple the number of experts in the tied
middle layers, leaving prelude and coda at the baseline expert count.
The $4{\times}$ variant returns each architecture's total parameter
count to within $1\%$ of its baseline (``iso-base''). Heterogeneous
width is implemented by constructing a temporary model with the
expanded expert count and transplanting its middle-layer MLPs into
the main model before applying expert tying.

\paragraph{Optimizer.} Identical optimiser to the ablation runs
(Appendix~\ref{app:repro-ablation}): Muon \cite{jordan2024muon, liu2025muon}
for $2$D hidden weights at $\eta_\text{Muon} = 2 \times 10^{-2}$ with
weight decay $0.1$, momentum $0.95$ Nesterov, $5$ Newton--Schulz
iterations; AdamW \cite{loshchilov2019adamw} optimizes the embeddings, output head, norm gains, biases, and routers at $\eta_\text{AdamW} = 0.1 \cdot \eta_\text{Muon} = 2 \times 10^{-3}$ with $(\beta_1, \beta_2) = (0.9, 0.95)$. We apply a uniform weight decay of $0.01$ to all AdamW parameters. While some large-scale Muon-MoE recipes apply heavy regularization ($0.1$) to AdamW parameters \cite{team2025kimi}, we adopt the lighter $0.01$ standard from the foundational Muon baseline \cite{jordan2024muon}. For the routers specifically, this provides a gentle physical shrinkage that works in tandem with the auxiliary $z$-loss to prevent logit explosion and softmax saturation.

The disparity between our $\eta_\text{Muon} = 2 \times 10^{-2}$ and the
$\eta \approx 2 \times 10^{-4}$ reported by \cite{liu2025muon}
and Kimi~K2~\cite{team2025kimi} is an artifact of two distinct shape-factor
conventions, not a genuine difference in update magnitude. \cite{liu2025muon}
multiply the orthogonalised update by $0.2 \cdot \sqrt{\max(d_\text{in},
d_\text{out})}$, which cancels the $1/\sqrt{\max(d_\text{in}, d_\text{out})}$
per-entry RMS of the orthogonalised update and leaves a shape-independent
per-entry update RMS of $0.2\,\eta$. The original Jordan
recipe~\cite{jordan2024muon} we use applies $\sqrt{\max(1,
d_\text{out}/d_\text{in})} \approx 1$ for the near-square FFN matrices in
our architectures, leaving a per-entry update RMS of
$\eta_\text{Muon}/\sqrt{\max(d_\text{in}, d_\text{out})}$. For our largest
ablation matrices ($\max d \approx 1024$), this gives $\sim\!6 \times 10^{-4}$
per entry, against $4 \times 10^{-5}$ per entry for K2 --- a $\sim\!16\times$
gap that closely matches the $\sim\!18\times$ width ratio between K2's hidden
size ($7168$) and our hidden size in the larger experiment configs ($384$), as predicted by the
$1/\text{width}$ LR-transfer rule of muP~\cite{yang2021mup}. We therefore
operate in the same effective regime as production Muon-MoE recipes; the
two conventions are interchangeable up to a corresponding LR rescaling.

\paragraph{3D expert tensors and tied-LR scaling.} The HuggingFace
MoE implementations stack expert weights into $3$D tensors of shape
$(E, 2 d_\text{ff,expert}, d_\text{model})$ (fused gate/up) and
$(E, d_\text{model}, d_\text{ff,expert})$ (down). Muon expects $2$D
inputs, so we register one $2$D proxy parameter per expert that shares
storage with its slice of the $3$D tensor; gradients are copied from
the $3$D parameter into the proxies via a post-backward hook before
the optimiser step. For \emph{tied} expert tensors, gradients accumulate
into the same parameter from $g$ layer use-sites, producing an effective
gradient of approximately $g$ times the untied magnitude. We compensate
by dividing the learning rate of tied-expert parameter groups by
$\sqrt{g}$ (\texttt{tied\_lr\_divisor=2.0} for $g{=}4$), as motivated
in Section~\ref{sec:opttied}; this is applied to both the Muon and
AdamW LR streams. Without this scaling the tied middle stack effectively
trains at a higher step size on its expert weights, which we found
degrades early-step stability and final loss in preliminary runs. The
non-tied baseline uses divisor $1$.

Decoupled optimizers apply weight decay as a penalty proportional to
the learning rate: $W \leftarrow W - \eta \nabla L - \eta \lambda W$.
Dividing $\eta$ by $\sqrt{g}$ for tied parameters therefore also reduces
their per-step regularization $\eta\lambda$ by the same factor. We leave
this uncompensated: all parameter groups, tied and untied, use the same
base $\lambda$ with no $\sqrt{g}$ adjustment, so tied parameters receive
proportionally less structural shrinkage per step. In an ablation at
$g{=}4$ we found that compensating the decay (multiplying $\lambda$ by
$\sqrt{g}$ for tied groups, restoring the untied $\eta\lambda$) gave
\emph{worse} final loss than leaving it uncompensated, so the simpler
uniform-$\lambda$ scheme is used throughout.

\paragraph{Learning rate schedule.} Cosine schedule with $100$-step
linear warmup, decaying from $\eta_\text{peak}$ to
$\eta_\text{min} = 0.1 \times \eta_\text{peak}$ at the final step;
Muon and AdamW peaks decay synchronously (and the tied-LR-divided
groups inherit the same shape).

\paragraph{Loss.} Standard next-token cross-entropy plus a load-balancing
auxiliary loss \cite{shazeer2017outrageously,fedus2022switch} with
coefficient $\alpha_\text{aux} = 10^{-2}$ and a router $z$-loss with
coefficient $\alpha_z = 10^{-4}$. Identical to the ablation runs of
Section~\ref{sec:component-ablation} (Appendix~\ref{app:repro-ablation}).

\paragraph{Training and data.} Training used $\approx\!10.5$B tokens total for the smaller configs, and $\approx\!15.7$B tokens for the 7B configs. This corresponds to $20{,}000$ or $30{,}000$ optimisation steps respectively. All configs use a global batch size of 256 sequences per step, or $524{,}288$ tokens, at 2048 sequence length. 
All small configurations use 16 micro batch size. For DDP=4 this means gradient accumulation~4. 
For additional single-GPU throughput experiments, the same micro-batch size results in grad accumulation 16.
Gradient clipping is active at a global norm $1.0$.
Mixed precision: bfloat16 forward and backward, float32 master
weights, \texttt{torch.compile} enabled where available. Larger configs use gradient checkpointing for saving GPU memory.
We use the same $75{:}25$ DCLM-edu~\cite{dclm-edu} and FinePhrase~\cite{finephrase} data mixture as the ablations
(Appendix~\ref{app:repro-ablation}), with the same data loader, tokenised with the same
\texttt{cl100k\_base} encoding.

\paragraph{Hardware.} Each run uses $4{\times}$H200 GPUs with PyTorch DDP. For the smaller config in Table \ref{tab:tiny_20k_final}, each $20{,}000$-step wall-clock time per run is roughly $5$~hours. The larger 7B scale experiments with $30{,}000$-step as in Table \ref{tab:fullscale} takes about 3.5 days ($g{=}4$ tied setting) on the 4 GPUs.

\paragraph{Efficiency \& Throughput.} 
To verify that expert tying does not introduce computational bottlenecks, we measured sustained training throughput across configurations at the $1{,}000$-step mark. Because hardware accelerators like the H200 are typically memory-bandwidth bound rather than compute bound in MoE architectures, the reduction in unique parameter count yields a direct wall-clock speedup. Our large untied OLMoE baseline of 7B-A1B, as detailed in Table \ref{tab:fullscale}, processes $41{,}859 $ tokens/sec. The $g{=}4$ tied topology (with experts tied) processes $51{,}777$ tokens/sec, a $23.7\%$ increase in throughput (identical global batch size, grad-accumulation 4x larger on baseline, as the identical local batch size setting results in \emph{out of memory}). On the smaller config shown in Table \ref{tab:tiny_20k_final}, the throughput gain is $15.7\%$ (again with DDP on 4 GPUs). For the small models we use identical global and local batch sizes in the comparison, as both can fit into GPU memory.
One should note that the throughput gains are also influenced by the reduced DDP communication need resulting from our tying. We relied on a fast intra-node GPU-to-GPU communication via NVLink via NVSwitch, 900 GB/s bidirectional bandwidth. In addition, another part of the gain (only used for 7B configs) is the possibility to run with 2x or 4x local batch size, due to the reduced GPU memory again from the parameter tying.

Finally, our code in plain PyTorch does not use any optimized compute kernels. Tailored compute kernels for tied layers could therefore likely result in further throughput improvements, both at training and inference time.
Our preliminary results confirm that the parameter savings translate directly into efficiency gains due to reduced memory traffic, reduced communication bandwidth, and improved~MFU.

\paragraph{Downstream evaluation.} We report macro-average $3$-shot
accuracy on \{ARC-Easy, ARC-Challenge, HellaSwag, PIQA, WinoGrande, OpenBookQA\}
via \texttt{lm-evaluation-harness}, evaluated on the final saved
checkpoint of each run.

\section{Monitoring Router Health and Expert Utilization}
\label{app:router-health}
A common failure mode in MoE training is router collapse, where the
gating network degenerates to selecting a small subset of experts and
the model effectively reduces to a dense network~\cite{shazeer2017outrageously}.
Because our architecture forces consecutive layers to share the same
underlying expert weights, it is important to verify that tying does
not encourage collapse or degrade routing diversity.
\paragraph{Logged metrics.} For every Section~\ref{sec:results} run
we track, averaged across all routing layers:
\begin{itemize}
\item \textbf{Auxiliary load-balancing loss} \cite{shazeer2017outrageously,fedus2022switch}
  ($\alpha_\text{aux} = 10^{-2}$), which penalises imbalanced expert usage.
\item \textbf{Mean routing entropy:} the Shannon entropy of the
  softmax-normalised routing distributions; low entropy indicates the
  router is committing to few experts.
\item \textbf{Router $z$-loss} \cite{zoph2022stmoe} ($\alpha_z = 10^{-4}$),
  which tracks the magnitude of the pre-softmax router logits and
  guards against logit explosion.
\item \textbf{Cross-loop agreement:} for tied configurations, the
  per-token top-$1$ routing agreement between layers that share an
  expert (the same metric defined in
  Section~\ref{sec:component-ablation}). High agreement indicates the
  shared expert is being reused with near-identical routing across
  loop positions; lower agreement indicates genuinely different
  per-layer use.
\end{itemize}
Note that the router parameters receive relatively weak weight decay in the AdamW split (see Appendix~\ref{app:repro-main}), so they undergo less structural
shrinkage. The $z$-loss prevents logit explosion in its absence.
\paragraph{Observations.}
Across all tying topologies the router remains healthy throughout
training: the auxiliary loss stays bounded (it does not run away or
collapse to zero), routing entropy stays well above zero, and the
$z$-loss keeps router logits in a stable range rather than diverging.
Tying does not induce router collapse at any group size.
\begin{figure}[ht]
\centering
\includegraphics[width=\linewidth]{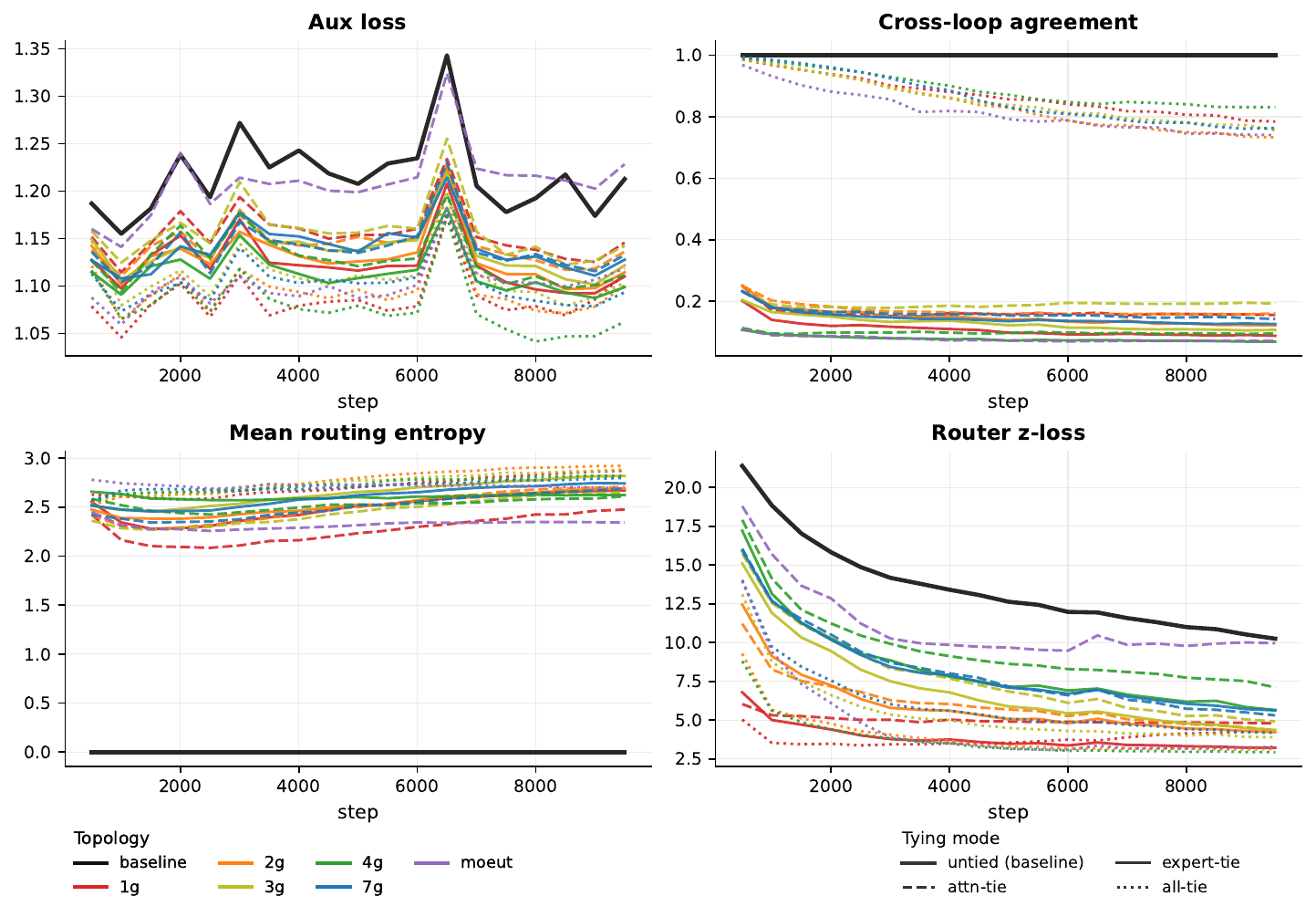}
\caption{Router-health metrics across the
Section~\ref{sec:component-ablation} fine-grained ablation runs, under
the $\sqrt{g}$ tied-LR scaling. \textbf{Colour} encodes topology
(baseline = black, $1g$ = red, $2g$ = orange, $3g$ = olive, $4g$ =
green, $7g$ = blue, MoEUT = purple); \textbf{line style} encodes tying
mode (\textsc{attn-tie} dashed, \textsc{expert-tie} solid,
\textsc{all-tie} dotted, untied baseline thick solid). The auxiliary
loss, routing entropy, and $z$-loss remain stable across all
configurations, indicating no router collapse under tying.
\textsc{expert-tie} tracks \textsc{attn-tie} closely on cross-loop
agreement and entropy---shared experts are reused with genuinely
different routing across loop positions---while \textsc{all-tie} shows
elevated cross-loop agreement, as expected when the router itself is
also shared.}
\label{fig:router-health-ablation}
\end{figure}

\end{document}